\documentclass[10pt,twocolumn,letterpaper]{article}

\usepackage{cvpr}              

\usepackage[accsupp]{axessibility}
\usepackage{graphicx}
\usepackage{amsmath}
\usepackage{amssymb}
\usepackage{bbm}
\usepackage{booktabs}
\usepackage{xfrac}
\usepackage{multirow}
\usepackage{pifont}
\usepackage{colortbl}

\usepackage{amsfonts}       
\usepackage{nicefrac}       
\usepackage{microtype}      
\usepackage{xcolor}         

\usepackage{breqn}
\usepackage{caption}
\usepackage{subcaption}
\usepackage{dsfont}
\usepackage{wasysym}

\usepackage{stfloats}
\usepackage{float}
\usepackage{makecell}
\usepackage{wrapfig,lipsum}
\usepackage{pythonhighlight}
\usepackage[pagebackref,breaklinks,colorlinks]{hyperref}
\usepackage[capitalize]{cleveref}
\crefname{section}{Sec.}{Secs.}
\Crefname{section}{Section}{Sections}
\Crefname{table}{Table}{Tables}
\crefname{table}{Tab.}{Tabs.}
\crefname{equation}{Eq.}{Eqs.}

\newcommand{\cmark}{\ding{51}}

\newcommand{\name}{GenVIS}

\def\cvprPaperID{3917} 
\def\confName{CVPR}
\def\confYear{2023}

\begin{document}

\title{A Generalized Framework for Video Instance Segmentation}

\author{
Miran Heo$^1${\qquad}Sukjun Hwang$^1${\qquad}Jeongseok Hyun$^1${\qquad}Hanjung Kim$^1${\qquad}\\
Seoung Wug Oh$^2${\qquad}Joon-Young Lee$^2${\qquad}Seon Joo Kim$^1$
\vspace{2mm}\\$^1$Yonsei University\qquad$^2$Adobe Research
\vspace{-2mm}
}

\maketitle

\begin{abstract}
   The handling of long videos with complex and occluded sequences has recently emerged as a new challenge in the video instance segmentation (VIS) community. 
   However, existing methods have limitations in addressing this challenge.
   We argue that the biggest bottleneck in current approaches is the discrepancy between training and inference.
   To effectively bridge this gap, we propose a \textbf{Gen}eralized framework for \textbf{VIS}, namely \textbf{GenVIS}, that achieves state-of-the-art performance on challenging benchmarks without designing complicated architectures or requiring extra post-processing.
   The key contribution of GenVIS is the learning strategy, which includes a query-based training pipeline for sequential learning with a novel target label assignment.
   Additionally, we introduce a memory that effectively acquires information from previous states.
   Thanks to the new perspective, which focuses on building relationships between separate frames or clips, GenVIS can be flexibly executed in both online and semi-online manner.
   We evaluate our approach on popular VIS benchmarks, achieving state-of-the-art results on YouTube-VIS 2019/2021/2022 and Occluded VIS (OVIS).
   Notably, we greatly outperform the state-of-the-art on the long VIS benchmark (OVIS), improving 5.6 AP with ResNet-50 backbone.
   Code is available at \url{https://github.com/miranheo/GenVIS}.
\end{abstract}

\section{Introduction}
\label{sec:intro}
\begin{figure}[t]
\begin{center}
\includegraphics[width=\linewidth]{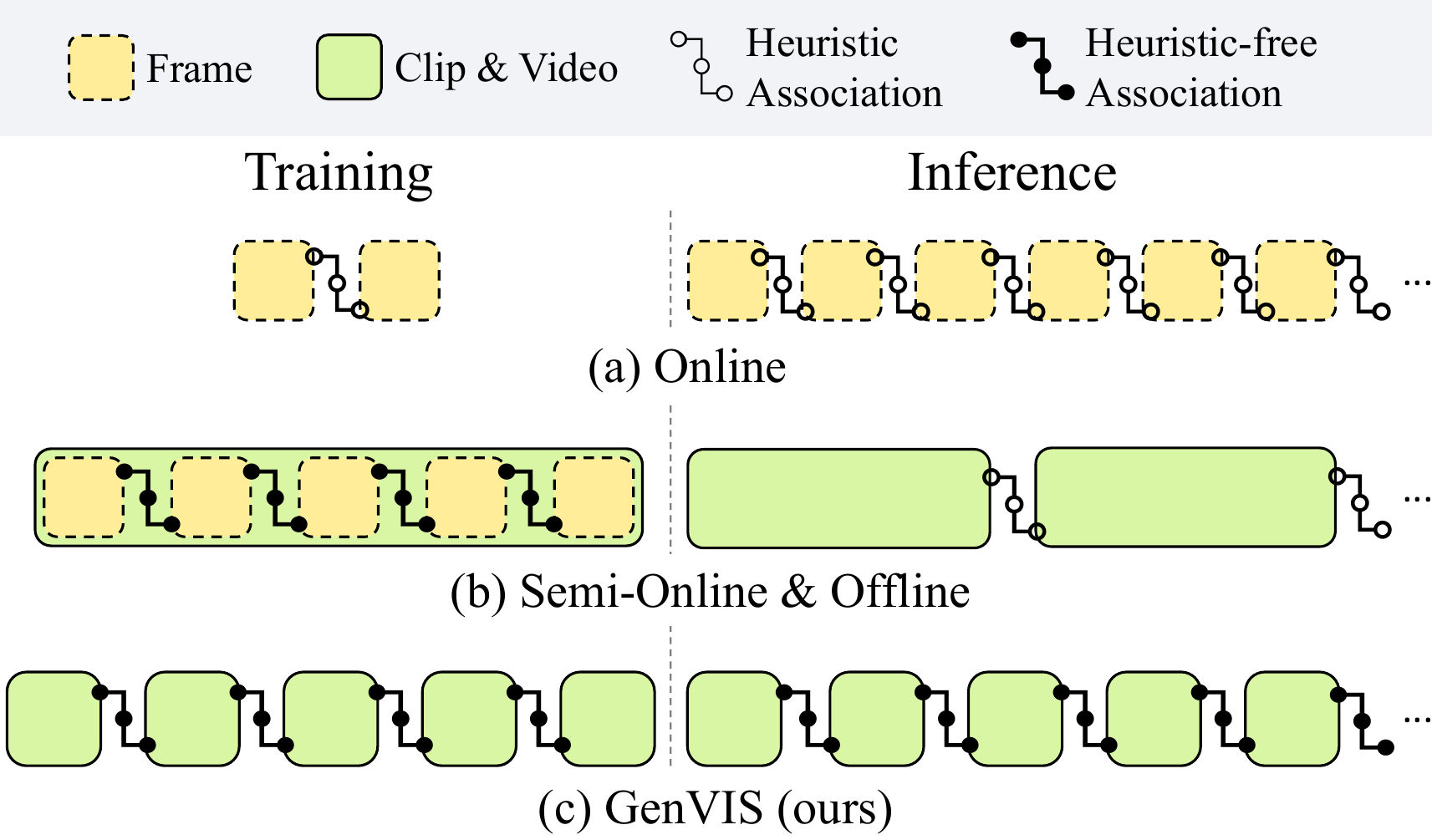}
\end{center}
\vspace{-6mm}
\caption{
Comparison between current VIS paradigms and our approach.
(a, b) While current methods use two separate paradigms based on the number of frames processed, we argue that the key challenge in processing real-world videos is building inter-clip assocications.
(b) Our proposed \textbf{GenVIS} addresses this challenge and can operate effectively in both online and semi-online manner without requiring hand-crafted post-processing.
}
\vspace{-5mm}
\label{fig:teaser}
\end{figure}
Video Instance Segmentation (VIS) is the task of identifying, segmenting, and tracking all objects in videos simultaneously.
With the emergence of datasets containing long and complex sequences, the research community is taking a step towards real-world applications.
While many papers have proposed solutions, the most notable performance improvement has been achieved by recent online methods using image-based backbones~\cite{MinVIS, IDOL}.
These results challenge the common belief that end-to-end semi-online or offline approaches (\emph{i.e.}, \cite{VisTR, IFC, Mask2Former-VIS, TeViT, SeqFormer, VITA}) trained on longer video clips would better model long-range object relationships.

We hypothesize that reason behind this somewhat surprising result is the presence or absence of an object association scheme between frames or clips that can scale to long videos.
Recent VIS methods, regardless of the approach, are driven by powerful image-level detectors~\cite{QueryInst, Mask2Former}, so detection and segmentation quality are already robust and comparable to each other.
To operate robustly in long videos, what the VIS really needs to focus on is the long-range tracking quality.
Although semi-online and offline methods are suitable for tracking objects within clips, they need to associate objects between clips to infer long videos, which is usually achieved using simple heuristics such as IoU matching~\cite{MaskProp, IFC}.

While online methods are more robust than semi-online/offline VIS solutions in processing long videos, we still see a significant room for improvement in their tracking approach. 
These methods only consider local contexts between adjacent frames during training, while test videos can exceed hundreds of frames~\cite{OVIS-Dataset}.
We believe that there is a better way to learn long-range temporal modeling that could fundamentally change the current VIS landscape. 


In this paper, we argue that the biggest bottleneck in handling long videos is the discrepancy between the training and inference scenarios.
Regardless of the previously defined paradigms (e.g., how many frames a method processes at once),
we need to focus on how to train the model. 
As illustrated in \cref{fig:teaser}, all previous methodologies use only a few frames or clips (\emph{e.g.}, one or two) for training, while real-world videos can have an unlimited length.
Therefore, they have to handle typical long-range tracking scenarios (\emph{e.g.}, newborn objects and re-identification) as exceptions through heuristics~\cite{IDOL}.


We introduce a \underline{\textbf{Gen}}eralized \underline{\textbf{VIS}} framework, namely \textbf{\name{}} (\cref{fig:teaser} (c)), that is designed to \emph{minimize} the gap between training and inference of long videos.
We take an existing offline VIS model (VITA~\cite{VITA}) as the backbone and applay a query-propagation method~\cite{EfficientVIS} for object association between clips. 
The essence of GenVIS is the novel training strategy.
By improving the training strategy of the base model, we achieve significant gains of 5.1 AP (Occluded VIS~\cite{OVIS-Dataset}) and 5.8 AP (YouTube-VIS 2022 Long Videos~\cite{MaskTrackRCNN}) in long and challenging benchmarks, outperforming all the previous methods by a large margin.

Our first proposal for improvement is to load multiple clips during training.
Unlike previous semi-online/offline VIS methods~\cite{MaskProp, VisTR, IFC, SeqFormer, Mask2Former-VIS} that focus on placing multiple frames in a single clip to strengthen intra-clip tracking, we propose to prioritize inter-clip tracking by learning the temporal relationship through multiple consecutive clips.
We believe that strong inter-clip reasoning is crucial for processing long videos in the real world.
As videos must be split into multiple clips that fit into GPU memory when processing long videos, inter-clip association is inevitable.
In this work, we use relatively short clip lengths (\emph{e.g.}, 1 to 7), but load as many clips as possible (\emph{e.g.}, usually more than 5).

More importantly, we propose a new learning criterion that enables seamless association through multiple consecutive clips.
Since we now deploy a sufficient number of clips, we can effectively simulate various inference scenarios at training time, covering newborn objects and objects that disappear and reappear.
Specifically, we propose the Unified Video Label Assignment (UVLA) that allows unique object queries to detect newly-appeared object and to keep them consistent once matching identities are obtained.
Our new learning criterion not only improves tracking performance but also removes all heuristics\footnote{Previously, \cite{EfficientVIS} need to determine whether a tracked query is valid or not with a confidence threshold.} required to handle new objects and re-identification from the inference stage.
In other words, our model infers videos exactly as it learned.
With these two proposals in the learning strategy, our base model outperforms all the previous methods on long video VIS benchmarks~\cite{OVIS-Dataset} without additional network modules. 

To further bridge the remaining gap between training and inference, we propose adopting a memory mechanism that stores previously decoded object queries. 
This mechanism is particularly useful for handling very long videos (or streaming video), where there is a limit to the number of clips that can be loaded at once.
To implement this, we add extra information for each object query by reading from its previous states. 
The memory mechanism results in meaningful improvements with only a small computational overhead.



Despite its simple framework, \name{} achieves state-of-the-art results on VIS benchmarks, outperforming previous methods on challenging long and complex video datasets (Occluded VIS~\cite{OVIS-Dataset} and YouTube-VIS 2022~\cite{MaskTrackRCNN}).
Our method also demonstrates strong generalization capability under online and semi-online settings\footnote{The term `general' describes how our proposed framework is designed to be versatile and able to operate in both online and semi-online manner. This flexibility is achieved by adjusting the clip length to 1, which allows us to convert the semi-online model into an online (per-frame) VIS model.}.
We provide additional analysis of the training and inference settings, which can be useful for balancing accuracy and efficiency tradeoffs.

\section{Related Works}
\label{sec:related}
\subsection{Video Instance Segmentation}
We categorize previous studies into two paradigms: 1) \emph{online}, and 2) \emph{semi-online \& offline}.
In this paper, we integrate the offline methods with the semi-online paradigm.

\textbf{Online} methods have been making considerable progress through advances in image-level object detection algorithms. 
MaskTrack R-CNN~\cite{MaskTrackRCNN} was the first attempt on the VIS task that puts its basis on the image instance segmentation model~\cite{MaskRCNN}.
For the extension to the video domain, MaskTrack R-CNN~\cite{MaskTrackRCNN} and follow-up works~\cite{SipMask, CrossVIS, SGNet} predict frame-independent outputs and make association using post-processing during the inference stage.
Recently, MinVIS~\cite{MinVIS} suggested that a strong, query-based image instance segmentation model~\cite{Mask2Former} inherently embeds distinctions of objects, and demonstrated competitive performance without video-based training.
Based on Deformable DETR~\cite{Deformable-DETR}, IDOL~\cite{IDOL} added a contrastive head that learns discriminative instance features between paired frames.

Inspired by Video Object Segmentation (VOS) methods~\cite{STM}, some previous works~\cite{VISOLO, PCAN, CompFeat, propose-reduce} adopted propagation approach. 
VISOLO~\cite{VISOLO} and PCAN~\cite{PCAN} integrated intermediate predictions and features to utilize the concept of memory~\cite{STM} for better performance.

\textbf{Semi-online \& Offline}
paradigms leverage multiple frames to take advantage of the rich temporal context. 
VisTR~\cite{VisTR} adopted~\cite{DETR, Transformer} and introduced the first end-to-end VIS model by taking a full video as an input and yielding video-level mask trajectories.
IFC~\cite{IFC} devised inter-frame communication that alleviates the heavy computation of VisTR and effectively encodes clip-wise information.
SeqFormer~\cite{SeqFormer} proposed an architecture that aggregates spatio-temporal contexts, iteratively decomposing multi-level features of clips using object queries.
VITA~\cite{VITA} introduced a new offline paradigm, showing that  video-level scene understanding can be achieved by building temporal interaction between frame-level object queries.

Several methods~\cite{MaskProp, IFC, TubeFormerDeepLab} take clip-level input and run sequentially with association algorithm during post-processing.
However, since the emergence of long video benchmarks\cite{OVIS-Dataset}, most existing offline methods~\cite{VisTR, IFC, Mask2Former-VIS, SeqFormer} cannot handle these benchmarks in an end-to-end manner as GPU memories cannot hold more than hundreds of frames.
Although IFC~\cite{IFC} proposed a post-processing algorithm to match clip-wise outputs, it depends on IoU between mask predictions of intersecting frames, thus computationally heavy.
EfficientVIS~\cite{EfficientVIS} proposed a method that stitches clip-wise outputs with minimal heuristics by making the model itself learn to infer over sequential clips.

\begin{figure*}[t]
\begin{center}
\includegraphics[width=\linewidth]{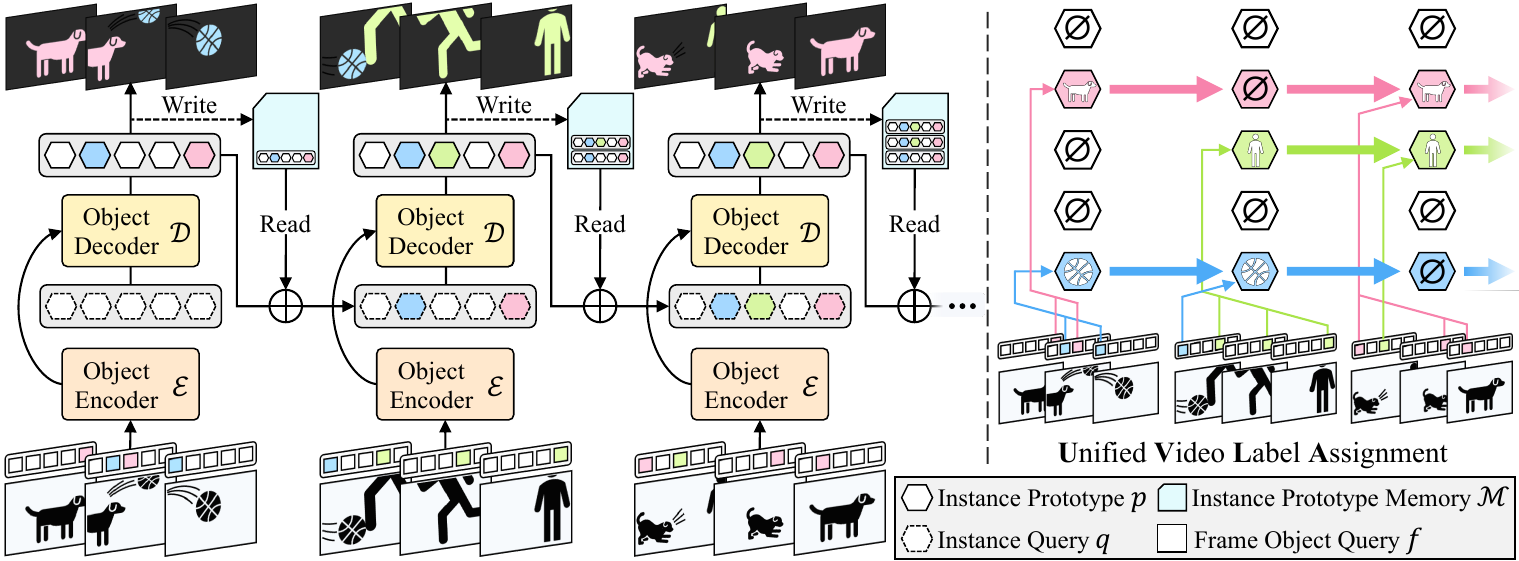}
\end{center}
\vspace{-4mm}
\caption{
(Left) Overview of our framework.
(Right) Unified Video Label Assignment.
By presenting this simple framework with novel ground-truth assignment strategy that resembles the inference pipeline, we significanly improve the accuracy on the VIS task without any bells-and-whistles.
$\varnothing$ : indicates ``no object'' as not matched to a ground-truth.
}
\vspace{-4mm}
\label{fig:main}
\end{figure*}
\subsection{Multi-Object Tracking}
\textbf{Query-based Trackers}~\cite{TrackFormer, MOTR, MeMOT} adopted query-based object detectors~\cite{DETR, Deformable-DETR} to track objects across frames with bounding boxes~\cite{SGT}.
TrackFormer~\cite{TrackFormer} follows a DETR-like architecture where newly appeared objects in a video are detected with object queries, but track queries are used to re-identify target objects which were detected and tracked in previous frames.
MeMOT~\cite{MeMOT} and MOTR~\cite{MOTR} also utilized track queries, but they enhance the training approach compared to TrackFormer by incorporating multiple frames and introducing additional modules that make use of previous track queries.
Nonetheless, these models entail complex heuristics with several handcrafted threshold values since both object and track queries are responsible for detecting objects in a scene, leading to duplicate detections.

\textbf{Clip-based Tracker}~\cite{TrackByClip} proposed a novel approach to improve long-term association by reformulating the standard tracking-detection scheme as a clip-to-clip matching problem.
Their observations are in line with our own, which suggest that focusing on inter-clip association is important for effectively utilizing temporal information.
\section{Method}
\label{sec:method}

We propose a novel generalized video instance segmentation framework GenVIS that sequentially associate clip-wise predictions.
To this end, we first split an input RGB video into ${N_{v}}$ non-overlapping clips $\{\mathcal{V}^{i} \in \mathbb{R}^{N_{f} \times H \times W \times 3}\}_{i=1}^{N_{v}}$, where each clip $\mathcal{V}$ consists of $N_{f}$ consecutive frames.
Here, $N_{f}$ is an integer less than the length of the input video; if $N_{f}=1$, our model runs in an online manner, otherwise semi-online. 
Sequentially taking a clip, GenVIS comes out with clip-level output queries that are used for predicting mask tracklets with a class probability for the input clip.
Then, the clip-level output queries from the current clip again become input object queries for the next clip.
Attributed to our newly defined label assignment strategy UVLA, GenVIS can 1) incorporate multiple clips (\emph{e.g.,} five clips) during training, which leads to better modeling real-video characteristics, and 2) narrow the gap between training and inference as the inter-clip associations can also be trained.
We also integrate information from past historical object queries by adopting a memory mechanism, which further improves the prediction qualities.
Finally, GenVIS presents \emph{significant accuracy} while completely \emph{removing heuristic matching algorithms}.

\subsection{Instance Prototypes}
As suggested in recent studies~\cite{MinVIS, VITA}, we hypothesize that a unique object is temporally coherent over a short period of time.
Therefore, an object can be compressed into a concise representation, \emph{instance prototype}.
Simply, instance prototypes are identical to output object queries that are used for predicting masks and categories in successful Transformer-based~\cite{Transformer} offline VIS architectures~\cite{IFC, Mask2Former-VIS, SeqFormer}.
However, such architectures have an underlying problem that only a limited number of clips can be loaded at training time, because they refer to dense spatio-temporal features to decode object queries.
For example, Mask2Former-VIS~\cite{Mask2Former-VIS} requires high-end GPUs with large memory capacity; in a single NVIDIA V100 GPU, 2 clips whose length is 2 frames (4 frames in total) are loaded for training.

To resolve the limitation, we adopt recently proposed VITA~\cite{VITA} (\cref{fig:main} (left)), which suggests that a video can be sufficiently represented with a collection of object-centric vectors.
Given a clip $\mathcal{V}$, we discard all features but the collection of $N_f \times N_q$ frame object queries ($f$) from a per-frame detector~\cite{Mask2Former}.
After building relations within $f$ using \emph{Object Encoder} ($\mathcal{E}$), we aggregate temporal contexts by inserting $N_q$ instance queries ($q$) into \emph{Object Decoder} ($\mathcal{D}$) and obtain instance prototypes ($p$) as $p=\mathcal{D}(q, \mathcal{E}(f))$.
The instance prototypes are forwarded into prediction heads that output clip-wise mask and class predictions $\{\hat{y}_{k}\}^{N_{\emph{q}}}_{k=1}$.

As illustrated in~\cref{fig:main} (left), we only train the four components that are disjoint from the backbone features: $\mathcal{E}$, $\mathcal{D}$, $q$, and $p$.
This leads to efficient memory consumption; more than 30 clips of 3 frames long (${\sim}100$ frames) can be used for training even with a single RTX 3090 GPU. In the VIS benchmark datasets, videos are sampled at 6 FPS; thus, 100 frames represent a video clip of 17 seconds long.
We find this strategy of using numerous clips during training is essential in designing long-range relationships and modeling real-video characteristics.
More training details are discussed in \cref{sec:imple_detail}.



\subsection{GenVIS}
\label{sec:cas_queries}
After obtaining instance prototypes from each clip, the next challenge is: how to associate predictions from adjacent clips.
Previous studies~\cite{MaskProp, IFC, MinVIS} bipartitely match such separate predictions using scores gauged under customized measurements, and they show competitive performance in relatively monotonous videos.
However, we empirically observe that these matching methods are vulnerable to scenes where many objects with similar appearances have complex trajectories.



Beyond the score-based associations, we take a further step to design a non-heuristic association under the motivation of eliminating the barriers between training and inference.
This can be achieved by training the instance prototypes to build relations between consecutive clips.
Specifically, as the instance prototypes encapsulate rich spatio-temporal and object-centric information, we formulate instance prototypes $p$ of the previous clip to be instance queries $q$ for the current clip as:
\begin{equation}
    \label{eq:clip_prototype}
    q^{i}= p^{i-1} = \mathcal{D}(q^{i-1}, \mathcal{E}(f^{i-1})),
\end{equation}
where $i$ is the index of the current clip.
Then, we place a strong condition that forces a unique instance prototype to represent a unique instance throughout the whole video.
To this end, we devise a new ground-truth assignment during training to support the strong condition, as existing label assignment strategies are defined within a single clip~\cite{SeqFormer, IFC, Mask2Former-VIS, MinVIS, VITA} or two clips~\cite{EfficientVIS} only.



\paragraph{Unified Video Label Assignment.}
We define \textbf{U}nified \textbf{V}ideo \textbf{L}abel \textbf{A}ssignment (UVLA) with three goals: 1) being capable of incorporating an arbitrary number of clips during training, 2) supporting the reuse of instance prototype for the next clip (\cref{eq:clip_prototype}), and 3) completely removing heuristics at inference.
To note, there have been similar attempts~\cite{Transtrack, MOTR} of reusing instance prototypes as additional instance queries for different clips.
However, they still require hyper-parameters and policies to remove or merge queries at inference.

For the definition of UVLA, we introduce two states to instance queries, \emph{`occupied'} and \emph{`unoccupied'}.
As illustrated in~\cref{fig:main} (right), instance queries become occupied (colored hexagons) if they are previously matched to a certain ground-truth, and they cannot be matched to another ground-truth object throughout a whole video.
On the other hand, only unoccupied instance queries (non-colored hexagons) have the opportunity to be matched to new ground-truth objects that have not appeared in the past.

Formally, at $i^{\text{th}}$ clip $\mathcal{V}^{i}$, we conduct one-to-one bipartite matching between $\{\hat{y}^{i}_{k}\}^{N_{\emph{q}}}_{k=1}$ and $\{y^{i}_{k}\}^{K^{i}_{\emph{new}}}_{k=1}$, where the former is the fixed-size set of $N_q$ predictions and the latter is the set of $K^{i}_{\emph{new}}$ ground-truths that are newly appeared in $\mathcal{V}^{i}$, respectively.
Following query-based detectors~\cite{DETR, Mask2Former}, we use Hungarian algorithm~\cite{Hungarian} to obtain the optimal assignment $\hat{\sigma}^{i}$ among a permutation of $N_q$ elements $\sigma \in \mathfrak{S}_{N_{\emph{q}}}$ as:
\begin{gather}
    \hat{\sigma}^{i} = \mathop{\arg\min}_{\sigma \in \mathfrak{S}_{N_{\emph{q}}}} \sum^{K^{i}_{\emph{new}}}_{k=1} \left(\mathcal{L}_{\emph{match}}(y^{{i}}_{k}, \hat{y}^{i}_{\sigma(k)}) + \alpha\cdot\mathds{1}_{\sigma(k) \in \Omega_{i-1}}\right), \raisetag{35pt}
\end{gather}
where $\mathcal{L}_{\emph{match}}$ is the pair-wise matching cost defined in VITA~\cite{VITA}, $\alpha$ is a large constant (\emph{e.g.,} $10^5$), and $\Omega_{i-1}$ is the indices of previously occupied instance queries.
The $|\Omega^{i-1}|$ occupied instance prototypes are directly matched to their previously paired ground-truths, and we can avoid these occupied instance prototypes being matched to newly added ground-truth by simply adding $\alpha\cdot\mathds{1}_{\sigma(k) \in \Omega_{i-1}}$ to the matching cost.
Then, we update $\Omega$ with the indices of newly matched queries as follows:
\begin{equation}
    \Omega^{i} = \Omega^{i-1} \cup \{\hat{\sigma}^{i}(k)\}^{K^{\emph{new}}_i}_{k=1},\qquad \Omega_{0} = \emptyset.
\end{equation}
The rest $(N_{q}-{K^{\emph{new}}_i}-|\Omega^{i-1}|)$ unoccupied predictions are matched to ``no object''.

With the proposed UVLA label assignment method, we can effectively narrow the gap between training and inference.
Concretely, we can use any number of clips (\emph{e.g.}, five) for training, allowing \name{} to learn real-video characteristics that appear in long sequences.
Furthermore, we can infer a video without any heuristics: if an object is detected by $k^{\text{th}}$ instance prototype at a certain clip, the $k^{\text{th}}$ instance query tracks the object until the input video ends.


\subsection{Instance Prototype Memory}
\label{sec:memory}
We further adopt memory~\cite{STM, XMem} to boost our long-range association ability by referring to information from past clips.
From the definition of UVLA, the occupied instance prototypes must be unceasingly matched to their designated ground-truths until the input video ends.
However, this policy would be difficult to be maintained when inferring an extremely long video, because the past information gradually fades out as sequentially processing clips.
For example, if an object appeared in the remote past and then disappeared for a long term, the occupied instance query may fail to recapture the same object.
To alleviate the limitation, we save concise information -- instance prototypes -- into the memory for each input clip and read them for improving predictions in the current clip.

Given the $i^{\text{th}}$ clip, all instance prototypes from past $i-1$ clips are gathered into the memory $\mathcal{M} = \{\{p^{k}_{j}\}^{N_q}_{j=1}\}^{i-1}_{k=1}$.
Using $q^{i}$ as a query to decode $\mathcal{M}$, we follow standard cross-attention~\cite{Transformer} and obtain an output $z^{i}$ that holds past information.
Lastly, we simply extend \cref{eq:clip_prototype} to incorporate $z^{i}$ for initializing object queries as $q^{i}=p^{i-1} + z^{i}$.
To note, for a query of index $j$, we narrow the scope to decode only the memory of same indices $\{p^{k}_{j}\}^{i-1}_{k=1}$.
Compared to globally decoding the memory, the index-wise decoding leads to better memory-reading efficiency and also improves accuracy.

\section{Experiments}
\label{sec:experiments}
\begin{table*}
\centering
{ 
\begin{tabular}{@{}c|lc|l|ccccc|ccccc@{}}
\toprule
\multicolumn{3}{l|}{\multirow{2}{*}{Method}}                        & \multirow{2}{*}{Backbone} & \multicolumn{5}{c|}{YouTube-VIS 2019} & \multicolumn{5}{c}{YouTube-VIS 2021}\\
\multicolumn{3}{l|}{}                                               &         & AP        & AP$_{50}$ & AP$_{75}$ & AR$_1$    & AR$_{10}$ & AP        & AP$_{50}$ & AP$_{75}$ & AR$_1$    & AR$_{10}$ \\
    \midrule
    \midrule
    
    \multirow{12}{*}{\rotatebox{90}{Semi-Online / Offline}}
    
    & \multicolumn{2}{l|}{EfficientVIS~\cite{EfficientVIS}}         & ResNet-50     & 37.9      & 59.7      & 43.0      & 40.3      & 46.6 
                                                                    & 34.0      & 57.5      & 37.3      & 33.8      & 42.5  \\
    & \multicolumn{2}{l|}{IFC~\cite{IFC}}                           & ResNet-50     & 41.2      & 65.1      & 44.6      & 42.3      & 49.6                                                                   & 35.2      & 55.9      & 37.7      & 32.6      & 42.9  \\
    & \multicolumn{2}{l|}{Mask2Former-VIS~\cite{Mask2Former-VIS}}   & ResNet-50     & 46.4      & 68.0      & 50.0      & -         & -  & 40.6      & 60.9      & 41.8      & -         & -     \\
    & \multicolumn{2}{l|}{TeViT~\cite{TeViT}}                       & MsgShifT      & 46.6      & 71.3      & 51.6      & 44.9      & 54.3   & 37.9      & 61.2      & 42.1      & 35.1      & 44.6  \\
    & \multicolumn{2}{l|}{SeqFormer~\cite{SeqFormer}}               & ResNet-50     & 47.4      & 69.8      & 51.8      & 45.5      & 54.8                                                                   & 40.5      & 62.4      & 43.7      & 36.1      & 48.1  \\
   
    & \multicolumn{2}{l|}{VITA~\cite{VITA}}                         & ResNet-50     & 49.8      & \textbf{72.6}      & 54.5      & 49.4      & \textbf{61.0}      
                                                                                    & 45.7      & \textbf{67.4}      & 49.5      & \textbf{40.9}      & \textbf{53.6}  \\
    & \multicolumn{2}{l|}{\textbf{\name{}$_{\text{semi-online}}$}}           & ResNet-50     & \textbf{51.3}  & 72.0    & \textbf{57.8}  & \textbf{49.5}  & 60.0  
                                                                                    & \textbf{46.3}  & 67.0 & \textbf{50.2} & 40.6 & 53.2 \\                  
    \cmidrule{2-14}
    
    & \multicolumn{2}{l|}{SeqFormer~\cite{SeqFormer}}               & Swin-L        & 59.3      & 82.1      & 66.4      & 51.7      & 64.4  
                                                                    & 51.8      & 74.6      & 58.2      & 42.8      & 58.1  \\
    & \multicolumn{2}{l|}{Mask2Former-VIS~\cite{Mask2Former-VIS}}   & Swin-L        & 60.4      & 84.4      & 67.0      & -         & -         
                                                                & 52.6      & 76.4      & 57.2      & -         & -     \\
    & \multicolumn{2}{l|}{VITA~\cite{VITA}}                         & Swin-L        & 63.0      & \textbf{86.9}      & 67.9      & \textbf{56.3}      & 68.1
                                                                                    & 57.5      & 80.6   & 61.0      & 47.7   & 62.6  \\   
    & \multicolumn{2}{l|}{\textbf{\name{}$_{\text{semi-online}}$}}           & Swin-L        & \textbf{63.8}     & 85.7              & \textbf{68.5}     & \textbf{56.3}     & \textbf{68.4} 
                                                                                    & \textbf{60.1}     & \textbf{80.9}     & \textbf{66.5}     & \textbf{49.1}     & \textbf{64.7}\\ 
    \midrule
    \midrule
    \multirow{9}{*}{\rotatebox{90}{Online}}
    & \multicolumn{2}{l|}{CrossVIS~\cite{CrossVIS}}                 & ResNet-50     & 36.3      & 56.8      & 38.9      & 35.6      & 40.7      
                                                                                    & 34.2      & 54.4      & 37.9      & 30.4      & 38.2  \\
    & \multicolumn{2}{l|}{VISOLO~\cite{VISOLO}}                     & ResNet-50     & 38.6      & 56.3      & 43.7      & 35.7      & 42.5      
                                                                                    & 36.9      & 54.7      & 40.2      & 30.6      & 40.9  \\
    & \multicolumn{2}{l|}{MinVIS~\cite{MinVIS}}                     & ResNet-50     & 47.4      & 69.0      & 52.1      & 45.7      & 55.7      
                                                                                    & 44.2      & 66.0      & 48.1      & 39.2      & 51.7  \\
    & \multicolumn{2}{l|}{IDOL~\cite{IDOL}}                         & ResNet-50     & 49.5      & \textbf{74.0}      & 52.9      & 47.7      & 58.7   
                                                                                    & 43.9      & \textbf{68.0}      & 49.6      & 38.0      & 50.9  \\
    & \multicolumn{2}{l|}{\textbf{\name{}$_{\text{online}}$}}                & ResNet-50 & \textbf{50.0} & 71.5 & \textbf{54.6} & \textbf{49.5} & \textbf{59.7} 
                                                                                    & \textbf{47.1} & 67.5 & \textbf{51.5} & \textbf{41.6} & \textbf{54.7} \\
    
    \cmidrule{2-14}
    & \multicolumn{2}{l|}{MinVIS~\cite{MinVIS}}                     & Swin-L        & 61.6      & 83.3      & 68.6      & 54.8      & 66.6 
                                                                                    & 55.3      & 76.6      & 62.0      & 45.9      & 60.8  \\   
    & \multicolumn{2}{l|}{IDOL~\cite{IDOL}}                         & Swin-L        & \textbf{64.3}      & \textbf{87.5}      & \textbf{71.0}      & 55.6      & 69.1
                                                                                    & 56.1      & 80.8      & 63.5      &  45.0      & 60.1  \\
    & \multicolumn{2}{l|}{\textbf{\name{}$_{\text{online}}$}}                & Swin-L        & 64.0 & 84.9 & 68.3 & \textbf{56.1} & \textbf{69.4}  
                                                                                    & \textbf{59.6}  & \textbf{80.9} & \textbf{65.8} & \textbf{48.7} & \textbf{65.0} \\
    \bottomrule
    \end{tabular}
} 
\vspace{-2mm}
\caption{
Comparisons on \textbf{YouTube-VIS 2019 and 2021 validation} sets.
We group the results by online or not online methods, and then, lightweight (\emph{e.g.}, ResNet-50) or powerful (\emph{e.g.}, Swin-L) backbone networks.
For each group, we \textbf{bold} the best value in every metric.
}
\vspace{-3mm}
\label{tab:ytvis2019_2021}
\end{table*}

\subsection{Datasets}
\label{sec:datasets}


We evaluate our method on two benchmarks: YouTube-VIS (YTVIS)~\cite{MaskTrackRCNN} and Occluded VIS (OVIS)~\cite{OVIS-Dataset}.
The YTVIS dataset has three versions (2019, 2021, 2022).
All versions tackle segmenting objects of 40 predefined categories in a video.
The dataset has been updated to include more challenging videos which are long and contain complex trajectories.
While the YTVIS~2021 and 2022 datasets share the same training set, YTVIS~2022 introduces $71$ extra videos on top of the validation set of YTVIS~2021.
We report the accuracy of the YTVIS~2021 videos and newly added YTVIS~2022 long videos, respectively.

OVIS is another challenging VIS dataset that consists of objects of 25 categories and targets a difficult scenario of heavy occlusions between objects.
Also, the OVIS dataset is significantly longer than the YTVIS datasets;
for example, the longest video of OVIS is ${\sim}50$ seconds long while that of YTVIS~2022 is ${\sim}10$ seconds long. 

\subsection{Implementation Details}
\label{sec:imple_detail}


While we adopt the architecture of VITA~\cite{VITA}, we replace the window-based self-attention layers for object tokens with the global attention layers.
As data augmentation techniques, random flipping and cropping are selected.
Also, we generate pseudo videos using images of the COCO dataset~\cite{COCO} and jointly train the model with the VIS datasets following \cite{SeqFormer}.
One batch consists of 8 videos; 5 clips ($N_{v}^{\emph{train}}=5$) are sampled from each video, and the length of clip ($N_{f}^{\emph{train}}$) is $1$ and $3$ frames for training online and semi-online versions, respectively.
Since we freeze the frame-level detector and backbone network, only a small size of memory is required for training.
Although a large number of frames are used for training (\emph{e.g.}, 40), the online version of GenVIS with a ResNet-50 backbone can be trained with even a single RTX 3090 GPU.
Our implementation is based on the \texttt{detectron2}~\cite{Detectron2} framework and is attached in the supplementary material.
\begin{table}
\centering
\resizebox{\linewidth}{!}
{
\begin{tabular}{@{}c|ll|ccccc@{}}
\toprule
\multicolumn{3}{l|}{Method}     & AP        & AP$_{50}$ & AP$_{75}$ & AR$_{1}$  & AR$_{10}$ \\
\midrule
\midrule
\multirow{4}{*}{\rotatebox{90}{ResNet-50\space\space}}
& \multicolumn{2}{l|}{MinVIS~\cite{MinVIS}}         & 23.3      & 47.9      & 19.3      & 20.2      & 28.0 \\
& \multicolumn{2}{l|}{VITA~\cite{VITA}}             & 32.6      & 53.9      & 39.3      & 30.3      & \underline{42.6} \\
\cmidrule{2-8}
& \multicolumn{2}{l|}{\textbf{\name{}$_{\text{online}}$}} & \underline{37.5}      & \underline{61.6}      & 41.5      & 32.6      & 42.2 \\
& \multicolumn{2}{l|}{\textbf{\name{}$_{\text{semi-online}}$}} & 37.2      & 58.5      & \underline{42.9}      & \underline{33.2}      & 40.4 \\
\midrule
\multirow{4}{*}{\rotatebox{90}{Swin-L\space}}
& \multicolumn{2}{l|}{MinVIS$^\dagger$~\cite{MinVIS}}         & 33.1      & 54.8      & 33.7      & 29.5     & 36.6 \\
& \multicolumn{2}{l|}{VITA$^\dagger$~\cite{VITA}}             & 41.1      & 63.0      & 44.0      & 39.3     & 44.3 \\
\cmidrule{2-8}
& \multicolumn{2}{l|}{\textbf{\name{}$_{\text{online}}$}} & \textbf{45.1} & 69.1 & \textbf{47.3} & 39.8 & \textbf{48.5} \\
& \multicolumn{2}{l|}{\textbf{\name{}$_{\text{semi-online}}$ }} & 44.3 & \textbf{69.9} & 44.9 & \textbf{39.9} & 48.4 \\
\bottomrule
\end{tabular}
}
\vspace{-2mm}
\caption{
%
Comparisons on the newly added \textbf{YTVIS 2022 long videos}.
$\dagger$: Evaluated using official repositories as the scores are not specified in original papers.
\underline{Underline} and \textbf{bold} denote the highest accuracy using ResNet-50 and Swin-L, respectively.
}
\vspace{-3mm}
\label{tab:ytvis_2022}
\end{table}

\begin{table}
\centering
\resizebox{\linewidth}{!}
{
\begin{tabular}{@{}c|ll|ccccc@{}}
\toprule
\multicolumn{3}{l|}{Method} & AP        & AP$_{50}$     & AP$_{75}$     & AR$_{1}$      & AR$_{10}$ \\
\midrule
\midrule

\multirow{8}{*}{\rotatebox{90}{ResNet-50}}
& \multicolumn{2}{l|}{CrossVIS~\cite{CrossVIS}}     & 14.9      & 32.7          & 12.1          & 10.3          & 19.8  \\
& \multicolumn{2}{l|}{VISOLO~\cite{VISOLO}}         & 15.3      & 31.0          & 13.8          & 11.1          & 21.7  \\
& \multicolumn{2}{l|}{TeViT$^\dagger$~\cite{TeViT}} & 17.4      & 34.9          & 15.0          & 11.2          & 21.8  \\
& \multicolumn{2}{l|}{VITA~\cite{VITA}}             & 19.6      & 41.2          & 17.4          & 11.7          & 26.0  \\
& \multicolumn{2}{l|}{MinVIS~\cite{MinVIS}}         & 25.0      & 45.5          & 24.0          & 13.9          & 29.7  \\
& \multicolumn{2}{l|}{IDOL~\cite{IDOL}}             & 30.2      & 51.3          & 30.0          & 15.0          & 37.5  \\
\cmidrule{2-8}
& \multicolumn{2}{l|}{{\textbf{\name{}$_{\text{online}}$}}}  & \underline{35.8} & \underline{60.8}          & \underline{36.2}           & 16.3          & \underline{39.6}   \\ 
& \multicolumn{2}{l|}{{\textbf{\name{}$_{\text{semi-online}}$}}}                          & 34.5      & 59.4          & 35.0          & \underline{16.6}         & 38.3  \\ 

\midrule

\multirow{5}{*}{\rotatebox{90}{Swin-L}}
& \multicolumn{2}{l|}{VITA~\cite{VITA}}             & 27.7      & 51.9          & 24.9          & 14.9          & 33.0  \\
& \multicolumn{2}{l|}{MinVIS~\cite{MinVIS}}         & 39.4      & 61.5          & 41.3          & 18.1          & 43.3  \\
& \multicolumn{2}{l|}{IDOL~\cite{IDOL}}             & 42.6      & 65.7          & 45.2          & 17.9          & \textbf{49.6}  \\
\cmidrule{2-8}
& \multicolumn{2}{l|}{\textbf{{\name{}$_{\text{online}}$}}} & 45.2      & 69.1         & \textbf{48.4}          & \textbf{19.1}         & 48.6  \\
& \multicolumn{2}{l|}{{\textbf{\name{}$_{\text{semi-online}}$}}} & \textbf{45.4}      & \textbf{69.2}          & 47.8          & 18.9         & 49.0  \\
\bottomrule
\end{tabular}
}
\vspace{-2mm}
\caption{
Comparisons on \textbf{OVIS validation} set.
\underline{Underline} and \textbf{bold} denote the highest accuracy using ResNet-50 and Swin-L, respectively.
$\dagger$ denotes using MsgShifT~\cite{TeViT} backbone.
}
\vspace{-3mm}
\label{tab:ovis}
\end{table}


\subsection{Main Results}
We compare GenVIS with state-of-the-art methods using both lightweight and powerful backbones, \emph{e.g.}, ResNet-50~\cite{ResNet} and Swin-L~\cite{Swin}, on the VIS benchmarks: YTVIS 2019/2021/2022 and OVIS.
GenVIS shows competitive performance on all benchmarks, and especially, it largely outperforms existing methods in the challenging datasets: YTVIS 2021/2022 and OVIS. 

\paragraph{YouTube-VIS 2019 \& 2021.}

In~\cref{tab:ytvis2019_2021}, we compare GenVIS with state-of-the-art methods on the YTVIS 2019 and 2021 benchmarks~\cite{MaskTrackRCNN}.
On the YTVIS 2019 benchmark, GenVIS achieves the highest AP with ResNet-50, while its AP is marginally lower than IDOL~\cite{IDOL} with Swin-L.
On the other hand, on the YTVIS 2021 benchmark which consists of more difficult videos, GenVIS surpasses not only IDOL~\cite{IDOL} with Swin-L in AP by $3.5$ but also with ResNet-50 by $3.2$.

\begin{table*}
\centering
{

\begin{tabular}{cccc|ccccc|ccccc}
\toprule
\multirow{2}{*}{$N_v^{\emph{train}}$} & \multirow{2}{*}{CL} & \multirow{2}{*}{UVLA} & \multirow{2}{*}{IPM} & \multicolumn{5}{c|}{OVIS} & \multicolumn{5}{c}{YouTube-VIS 2022 Long Videos}\\
&  &  &  & AP                                & AP$_{50}$     & AP$_{75}$     & AR$_{1}$      & AR$_{10}$ & AP     & AP$_{50}$ & AP$_{75}$     & AR$_{1}$      & AR$_{10}$ \\
\midrule
\midrule
1 &   &  &  & 24.1 & 44.9 & 22.6 & 13.4 & 30.2
            & 25.4 & 43.6 & 26.8 & 22.7 & 29.6 \\
2 & \cmark &  &  & 27.4 & 47.2 & 27.4 & 13.9 & 32.6
            & 29.6 & 49.3 & 32.0 & 26.3 & 35.0 \\
\midrule
5 & \cmark &  &  & 29.8 & 50.4 & 31.3 & 15.0 & 34.8
            & 32.8 & 51.9 & 36.4 & 32.4 & 39.1 \\
5 & \cmark & \cmark &  & 32.5 & 56.5 & 33.2 & 16.1 & 36.9 
             & 35.4 & 53.4 & 42.8 & 32.1 & 38.6\\
5 & \cmark & \cmark & \cmark & 33.4  & 57.8 & 34.5 & 15.7 & 37.7 
              & 36.9 & 56.5 & 44.7 & 32.1 & 41.4 \\
\bottomrule
\end{tabular}
}
\vspace{-2mm}
\caption{
Ablation study of the methods for learning inter-clip association (CL and UVLA) and the memory module (IPM).
$N_v^{\emph{train}}$ is the number of clips used for training.
CL and UVLA denote correspondence learning and unified video label assignment, respectively.
}
\vspace{-3mm}
\label{tab:ablation_method}
\end{table*}

\paragraph{YouTube-VIS 2022.}
As shown in~\cref{tab:ytvis_2022}, we also present comparisons on the YouTube-VIS 2022 benchmark~\cite{MaskTrackRCNN} which is more challenging than the previous benchmarks (2019 and 2021).
Our \name{} shows significant improvement over the previous state-of-the-art methods: MinVIS~\cite{MinVIS} and VITA~\cite{VITA}.
Based on the strong ability to track objects even under complex trajectories, our method using a ResNet-50~\cite{ResNet} backbone achieves 37.5 AP.
This is the highest accuracy among the models that use the same backbone, and also higher than MinVIS using a Swin-L~\cite{Swin} backbone.
Attaching our method on top of a Swin-L backbone, GenVIS shows 45.1 AP which is 4.0 AP higher than VITA.
Although GenVIS$_{\text{semi-online}}$ shows slightly lower AP than its online version, it still outperforms MinVIS and VITA.

\paragraph{OVIS.}
We further validate the competitiveness of \name{} on the OVIS~\cite{OVIS} validation set as shown in~\cref{tab:ovis}.
Compared to IDOL~\cite{IDOL} which previously ranked the highest accuracy, our method achieves 5.6 and 2.8 AP improvements on top of ResNet-50 and Swin-L backbones, respectively.

\begin{table*}
\begin{minipage}{0.35\linewidth}
    \centering
    \vspace{0mm}
    {
        \begin{tabular}{c|cc}
        \toprule
            & Global & Index \\
        \midrule
        \midrule
        AP  & 31.8   & 33.4 \\
        \bottomrule
        \end{tabular}
    }
    \vspace{-2mm}
    \caption{
    Ablation study of memory decoding method in IPM on the OVIS validation set.
    }
    \vspace{3mm}
    \label{tab:memory_agg_IPM}

    \resizebox{1.0\linewidth}{!}
    {
        \begin{tabular}{c|ccccc}
        \toprule
        $N_{f}$  & AP        & AP$_{50}$     & AP$_{75}$     & AR$_{1}$      & AR$_{10}$ \\
        \midrule
        \midrule
        1            & 33.0 & 56.8 & 34.3 & 15.7 & 37.5 \\ 
        \textbf{3}   & \textbf{33.4} & \textbf{57.8} & \textbf{34.5} & \textbf{15.7} & \textbf{37.7}  \\
        5            & 31.3 & 56.0 & 32.8 & 15.3 & 35.9  \\
        7            & 29.7 & 51.4 & 30.7 & 14.6 & 34.0 \\
        \bottomrule
        \end{tabular}
    }
    \vspace{-2mm}
    \caption{
    Ablation study of the clip length ($N_{f}$) on the OVIS validation set.
    The same value ($N_{f}$) is used for training ($N_{f}^{\emph{train}}$) and inference ($N_{f}^{\emph{eval}}$).
    }
    \label{tab:window_size_train}
    \end{minipage}
    \hfill
    \begin{minipage}{0.63\linewidth}
    \resizebox{0.95\linewidth}{!}
    {
        \centering
        \includegraphics[width=1.\linewidth]{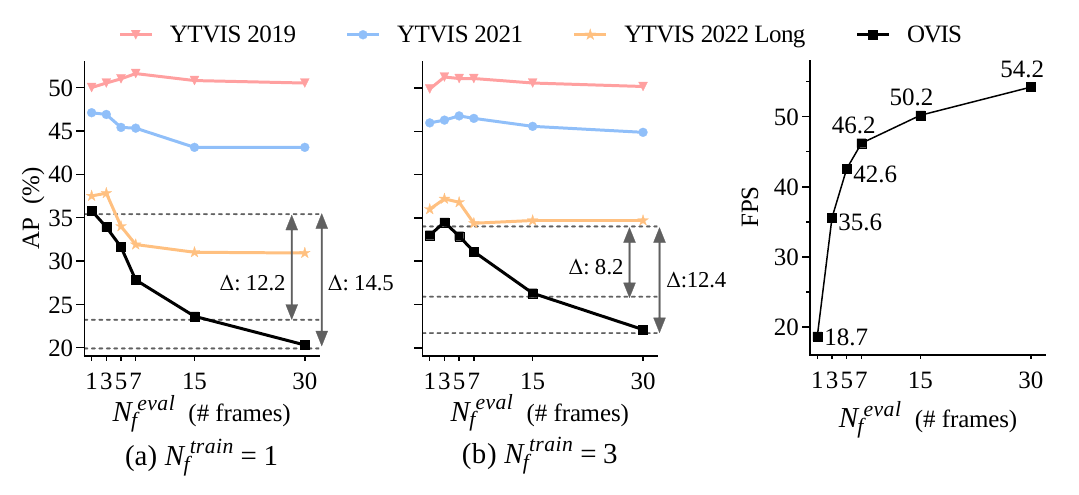}
    }
    \vspace{-2mm}
    \captionof{figure}{
    Deployment of GenVIS using the different lengths of a clip ($N_f^{\emph{eval}}$) for inference on the Youtube-VIS 2019/21/22 and OVIS validation sets. We validated two different GenVIS models trained using $N_f^{\emph{train}}$ of 1 and 3 which represent online and semi-online settings, respectively.
    FPS is measured on a single A100 GPU.
    The models trained for the main experiments are used.
    }
    \label{fig:window_size_eval}
\end{minipage}

\end{table*}

\subsection{Ablation Studies}
\label{sec:ablation}
We conduct ablation studies to verify the proposed learning method and memory module and show flexible usage of GenVIS in both online and semi-online settings. 
Here, we use a ResNet-50~\cite{ResNet} backbone and do not use pseudo videos generated from the COCO dataset for training.

\paragraph{Effect of the proposed learning method and memory}
We initiate the experiment with the VITA~\cite{VITA} model that regards an entire video as one long clip ($N_{v}=1$) and learns the direct association of objects within a clip.
Since we split a video into non-overlapping clips of three frames, we adopt the matching algorithm of MinVIS~\cite{MinVIS} that uses the cosine similarity of instance prototypes for the association across clips (inter-clip association).
This baseline shows AP of $24.1$ and $25.4$ in OVIS and YTVIS 2022 Long, respectively, as shown in \cref{tab:ablation_method}.
As the next baseline, we apply the correspondence learning~\cite{EfficientVIS} between two clips that the same query refers to the same object across clips.
This improves AP to $27.4$ and $29.6$, respectively.

On top of these two baselines, we extend learning of correspondence between two clips to five clips.
Such naive adaptation leads to improvement in both OVIS ($27.4 \rightarrow 29.8$) and YTVIS 2022 Long ($29.6 \rightarrow 32.8$).
This demonstrates that training with multiple clips allows the model to effectively learn the inter-clip association.
When we adopt UVLA, our label assignment strategy, carefully designed for learning correspondence between multiple clips by considering complex scenarios occurring in long videos, such as a temporary disappearance of objects, AP is significantly improved in both datasets ($29.8 \rightarrow 32.5$ and $32.8 \rightarrow 35.4$).
Incorporating the instance prototype memory (IPM) additionally increases AP.
In a nutshell, our proposed training method improves AP by $6.0$ and $7.3$ in OVIS and YTVIS 2022 Long, respectively, even though we maintain the simple and heuristic-free inter-clip association method.

\paragraph{Memory decoding in IPM.}
As shown by \cref{tab:memory_agg_IPM}, global decoding of instance prototypes in the memory does not show an improvement in the accuracy compared to index-wise decoding.
Also, index-wise decoding is computationally more scalable with a long video, thus, we decide to use it in IPM.

\vspace{-2mm}
\paragraph{Length of clip ($N_{f}$) for training.}
With the fixed number of clips ($N_{v}=5$), we experiment with the length of clip ($N_{f}$) used for training.
As shown in \cref{tab:window_size_train}, there is no notable change in AP when $N_{f}$ increases from 1 to 3.
However, higher $N_{f}$ values (\emph{e.g.}, 5 and 7) bring about degradation of AP, for example, a decrease of 2.1 and 3.7, respectively, compared to $N_{f}=3$.
Based on this empirical observation, we use $N_{f}=3$ for training the semi-online version of GenVIS.

\vspace{-2mm}
\paragraph{Flexibility of GenVIS for online and semi-online settings.}
As shown in \cref{fig:window_size_eval}, GenVIS is flexible to use different clip lengths for inference ($N_{f}^{\emph{eval}}$) although it is trained with a clip length ($N_{f}^{\emph{train}}$) of 1 or 3.
In the YTVIS 2019 dataset, increasing the window size from one to seven improves AP even though GenVIS is trained with a clip of one frame.
Since YTVIS 2021 \& 2022 and OVIS datasets have more complex trajectories of objects than YTVIS 2019, using a larger window size for inference leads to lower AP, especially in the OVIS dataset where objects are heavily occluded in each other.
In these datasets, using a short clip to extract the instance prototypes is more effective than a long clip which is also observed in \cref{tab:window_size_train}. 
In contrast, using a longer clip outputs temporally compact instance prototypes that lead to faster execution speed and memory efficiency.
In the case of such usage, it is better to use the GenVIS trained using a long clip (\emph{e.g.}, $N_{f}^{\emph{train}}=3$) since the decrease in AP is lower than the online version which is trained with $N_{f}^{\emph{train}}=1$.

\subsection{Qualitative Results}
\label{sec:qualitative}
\begin{figure*}[t]
\begin{center}
\includegraphics[width=\linewidth]{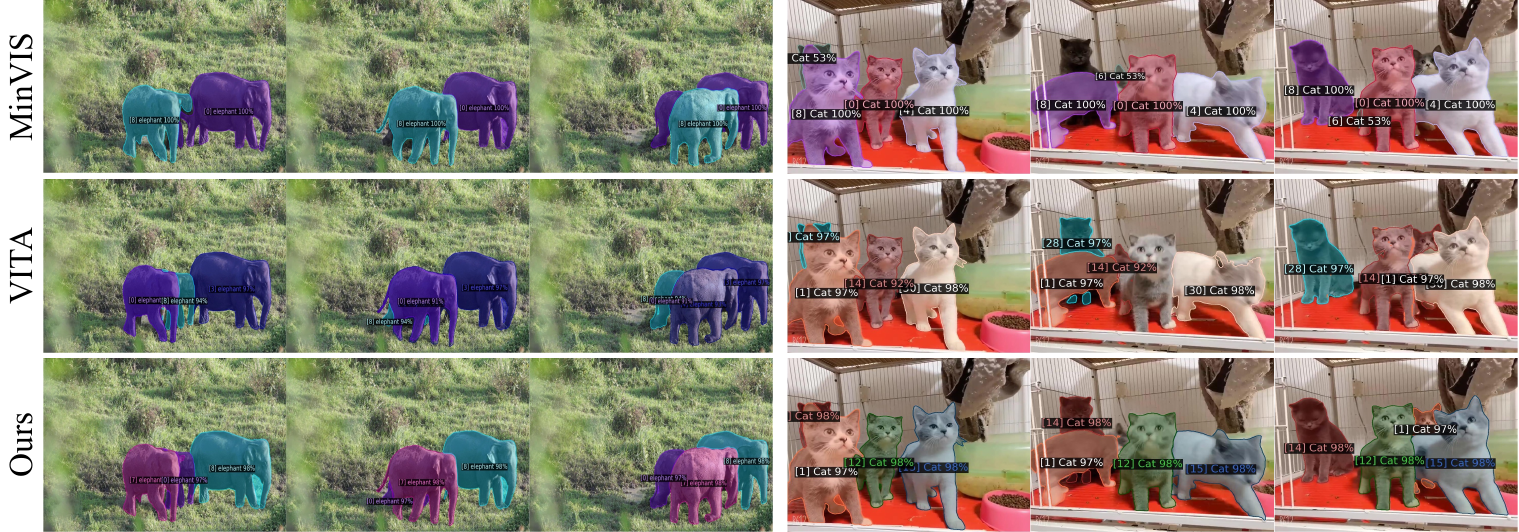}
\end{center}
\vspace{-4mm}
\caption{
Qualitative comparisons of our method, GenVIS, with the state-of-the-art methods: MinVIS~\cite{MinVIS} and VITA~\cite{VITA}.
The videos on the left and right are from YouTube-VIS 2022~\cite{MaskTrackRCNN} and OVIS~\cite{OVIS-Dataset} datasets, respectively.
GenVIS shows impressive accuracy in these complicated scenes where the objects look similar crossing each other. Objects with the same identity are displayed in the same color.
}
\vspace{-3mm}
\label{fig:qualitative}
\end{figure*}
\cref{fig:qualitative} shows the visualization results of our GenVIS and recent online~\cite{MinVIS} and offline VIS methods~\cite{VITA} on the benchmarks consisting of long and complicated scenarios.
In the selected scenes, the animals are similar in appearance, and their trajectories are too complex to keep tracking the same instance.
In the scene of YTVIS 2022, MinVIS fails to segment the elephant which is severely occluded by the elephant in front of the target, and VITA fails to track the front elephant resulting in the change of its ID.
However, GenVIS successfully segments and tracks all objects in this challenging scene.
GenVIS performs well even in the scene of OVIS where the left-front cat is moving behind two cats.

\section{Limitations and Future Works}
\label{sec:discussions}
Our method achieves state-of-the-art performance on long videos in both online and semi-online manners.
Also, we demonstrate the trade-offs between the accuracy and the efficiency for various learning options.
While the proposed framework provides a choice of the length of the clip without modifying the architecture, once the length of the clip is determined, our method treats an input video uniformly in that unit.
However, there can be a variety of situations in a single video.
For example, if a scene continues monotonically, it may be advantageous to process it broadly and put it in memory without taking the scope small, whereas frame-level processing could be beneficial in the opposite case.
Therefore, designing adjustable windows will be an interesting future direction.
In addition, because we take a label assignment strategy that gives unique IDs to unique object queries, the strategy can be limited if there are very large numbers of objects across the video.
Designing an algorithm integrated with memory could be another future direction.

\section{Conclusion}
\label{sec:conclusion}
In this paper, we propose \name{} a generalized framework for video instance segmentation (VIS).
To seamlessly bridge the gap between training and inference, we propose the following: 1) a training strategy that can involve multiple clips, 2) training the associations between separate clips with the novel label assignment - UVLA, and 3) the use of memory to alleviate vanishing information in long videos.
By integrating these proposals into a single framework, we demonstrate the effectiveness of our method by achieving state-of-the-art results on multiple benchmarks.

\vspace{3mm}
\section*{Acknowledgements}
\label{sec:acknowledgements}
This work was supported by Artificial Intelligence Graduate School Program under Grant 2020-0-01361, and by Institute of Information communications Technology Planning Evaluation (IITP) grant funded by the Korea government (MSIT) (No. 2022-0-00113, Developing a Sustainable Collaborative Multi-modal Lifelong Learning Framework, and No. 2014-3-00123, Development of High Performance Visual BigData Discovery Platform for LargeScale Realtime Data Analysis).


{
    \clearpage
    \small
    \bibliographystyle{plain}
    \bibliography{references}
}
\newpage
\end{document}